\pdfoutput=1

\documentclass[11pt]{article}

\usepackage{acl}

\usepackage{times}
\usepackage{latexsym}
\usepackage{multirow}
\usepackage{multicol}
\usepackage{booktabs}

\usepackage[T1]{fontenc}

\usepackage[utf8]{inputenc}

\usepackage[scaled=0.8]{beramono}

\usepackage{microtype}

\usepackage[normalem]{ulem}

\usepackage{tikz}
\usepackage{tikzscale}
\usepackage{pgfplots}
\usetikzlibrary{trees}

\title{Phylogeny-Inspired Adaptation of Multilingual Models to New Languages}

\author{Fahim Faisal, Antonios Anastasopoulos\\
Department of Computer Science, George Mason University\\
\texttt{\{ffaisal,antonis\}@gmu.edu}}

\begin{document}
\maketitle
\begin{abstract}
Large pretrained multilingual models, trained on dozens of languages, have delivered promising results due to cross-lingual learning capabilities on a variety of language tasks. Further adapting these models to specific languages, especially ones unseen during pre-training, is an important goal toward expanding the coverage of language technologies. 
In this study, we show how we can use language phylogenetic information to improve cross-lingual transfer leveraging closely related languages \textit{in a structured, linguistically-informed manner}. We perform adapter-based training on languages from diverse language families (Germanic, Uralic, Tupian, Uto-Aztecan) and evaluate on both syntactic and semantic tasks, obtaining more than 20\% relative performance improvements over strong commonly used baselines, especially on languages unseen during pre-training.
\footnote{Code and data are publicly available: \url{https://github.com/ffaisal93/adapt_lang_phylogeny}}
\end{abstract}

\section{Introduction}
Language models have now become the standard for building state-of-the-art Natural Language Processing (NLP) systems. 
Beyond monolingual models, large-scale multilingual models covering more than 100 languages are now available, such as \textsc{xlm-r} by \citet{conneau-etal-2020-unsupervised} and m\textsc{bert} by \citet{devlin-etal-2019-bert}, achieving competitive performance across languages from a variety of families and using various scripts.

Still, most of the 6500+ spoken languages in the world~\cite{hammarstrom2016linguistic}
are not covered --remaining unseen-- by those models. 
Even languages with millions of native speakers like Lingala (with 15-20 million speakers in central Africa, mostly D.R. Congo) or Bambara (spoken by around 5 million people in Mali and neighboring countries) are not covered by any available language models at the time of writing.

A recent line of work (see \S\ref{sec:background}) has shown that these large multilingual language models (MLMs) can be finetuned on individual languages to further improve performance. Even better, they can be even adapted to languages \textit{unseen} during the pre-training stage.\footnote{The potential of such approaches is conditioned on the language's script and data availability, of course.}

This work focuses on using adapters, a popular framework for such adaptation that has been proven successful for zero-shot and few-shot cross-lingual transfer. In particular, we significantly improve the adapter framework by drawing inspiration from a simple insight: that the adapters of related languages would likely need to perform the same function, and thus adapters could be trained leveraging multiple related languages. We impose a phylogenetically-inspired tree hierarchy for parameter-sharing between adapters and show empirically that our approach leads to large improvements with experiments on three NLP tasks on several language families.

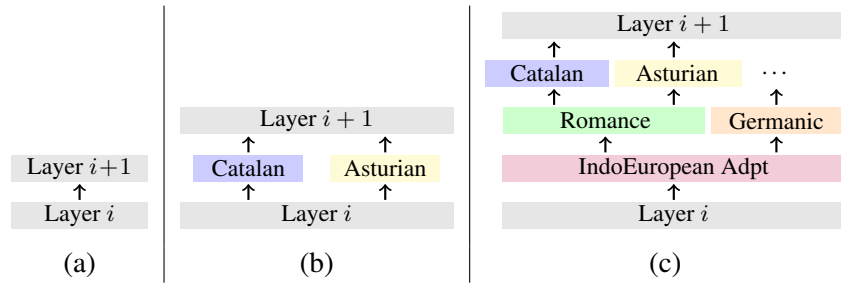
\begin{figure*}[t]
\tikzset{seq/.style={draw=none,fill=gray!20}}
\tikzset{seqblue/.style={draw=none,fill=blue!20}}
\tikzset{seqyellow/.style={draw=none,fill=yellow!20}}
\tikzset{seqgreen/.style={draw=none,fill=green!20}}
\tikzset{seqred/.style={draw=none,fill=red!20}}
\tikzset{seqpurple/.style={draw=none,fill=purple!20}}
\tikzset{seqorange/.style={draw=none,fill=orange!20}}
\tikzset{seqlabel/.style={font={\small}}}
\tikzset{layer/.style={->,thick}}
\centering
\begin{tabular}{c|c|c}
\begin{tikzpicture}[x=.9cm,y=0.7cm]
\draw[seq] (-1,-0.25) rectangle (1,0.25);
\node[seqlabel] at (0,0) {\small{Layer $i$}};
\draw[layer] (0,0.3) -- (0,0.6);
\draw[seq] (-1,0.65) rectangle (1,1.15);
\node[seqlabel] at (0,0.9) {\small{Layer $i\!+\!1$}};
\end{tikzpicture}
&
\begin{tikzpicture}[x=.9cm,y=0.7cm]
\draw[seq] (-2,-0.25) rectangle (2,0.25);
\node[seqlabel] at (0,0) {\small{Layer $i$}};
\draw[layer] (-1,0.3) -- (-1,0.6);
\draw[seqblue] (-1.8,0.65) rectangle (-0.2,1.15);
\node[seqlabel] at (-1,0.9) {\small{Catalan}};
\draw[layer] (1,0.3) -- (1,0.6);
\draw[seqyellow] (0.2,0.65) rectangle (1.8,1.15);
\node[seqlabel] at (1,0.9) {\small{Asturian}};
\draw[layer] (-1,1.2) -- (-1,1.5);
\draw[layer] (1,1.2) -- (1,1.5);
\draw[seq] (-2,1.55) rectangle (2,2.05);
\node[seqlabel] at (0,1.8) {\small{Layer $i+1$}};
\end{tikzpicture}
&
\begin{tikzpicture}[x=.9cm,y=0.7cm]
\draw[seq] (-4,-0.25) rectangle (1,0.25);
\node[seqlabel] at (-1.5,0) {\small{Layer $i$}};

\draw[layer] (-1.5,0.3) -- (-1.5,0.6);
\draw[seqpurple] (-4,0.65) rectangle (1,1.15);
\node[seqlabel] at (-1.5,0.9) {\small{IndoEuropean Adpt}};


\draw[layer] (-2.5,1.2) -- (-2.5,1.5);
\draw[seqgreen] (-4,1.55) rectangle (-1.05,2.05);
\node[seqlabel] at (-2.5,1.8) {\small{Romance}};

\draw[layer] (0,1.2) -- (0,1.5);
\draw[seqorange] (-0.95,1.55) rectangle (1,2.05);
\node[seqlabel] at (0,1.8) {\small{Germanic}};
\draw[layer] (0,2.1) -- (0,2.4);
\node[seqlabel] at (0,2.7) {\ldots};

\draw[layer] (-3.25,2.1) -- (-3.25,2.4);
\draw[seqblue] (-4.25,2.45) rectangle (-2.45,2.95);
\node[seqlabel] at (-3.35,2.7) {\small{Catalan}};

\draw[layer] (-1.5,2.1) -- (-1.5,2.4);
\draw[seqyellow] (-2.35,2.45) rectangle (-0.5,2.95);
\node[seqlabel] at (-1.45,2.7) {\small{Asturian}};

\draw[layer] (-1.5,3.0) -- (-1.5,3.3);
\draw[layer] (-3.25,3.) -- (-3.25,3.3);

\draw[seq] (-4,3.35) rectangle (1,3.85);
\node[seqlabel] at (-1.5,3.6) {\small{Layer $i+1$}};
\end{tikzpicture}\\
(a) & (b) & (c) 
\end{tabular}
    \caption{Incorporating phylogeny into neural models with adapters: 
    starting with an unadapted model (a), current practice uses language-specific adapters between layers (b). We instead 
    impose a phylogeny-informed tree hierarchy over adapters as in (c).}
    \label{fig:phylogeny}
\end{figure*}
 
\section{Background}
\label{sec:background}

\paragraph{Adapting Large-Scale Models to Low-Resource Languages} Multilingual language models (MLMs) can be used directly on unseen languages, or they can also be adapted using unsupervised methods. For example, \citet{han-eisenstein-2019-unsupervised} successfully used continued training with masked language modeling on unlabeled data to adapt an English BERT model to Early Modern English for sequence labeling. More recently, \citet{muller-etal-2021-unseen} employed the same strategy (enhanced with transliteration to handle languages with different scripts) to adapt models for several unseen-during-pretraining languages.

\paragraph{Adapter Units} Instead of fine-tuning the whole model, a more promising approach for adaptation uses dedicated units (\textit{adapter units}) that are injected between the layers of the pre-trained model (see example in Figure~\ref{fig:phylogeny}.b) and can be trained on a new language, domain, or task~\cite{vilar-2018-learning,houlsby2019parameter,pfeiffer-etal-2020-adapterhub,pfeiffer-etal-2020-mad}. There are two advantages in fine-tuning only these adapter components. Since they consist of only a small number of parameters, they can be adequately trained with a small number of training examples. In addition, as the pre-trained model remains invariant, they render \textit{catastrophic forgetting}~\cite{french1999catastrophic,Kirkpatrick3521} a non-issue. 

Nevertheless, the application of these adapters has so far followed a simple, straight-forward protocol: insert the adapters, and train them individually for a new task or language. 
In our work, we investigate how we can improve this process, by incorporating additional linguistic information. The core idea is to incorporate phylogenetic information in the adapters' organization.

\section{Phylogeny-Inspired Adaptation}
\label{sec:phylogeny}

\paragraph{Motivation} Intuitively, given the similarities between two related lects (e.g. Catalan and Asturian), one should exploit that relationship to inform the adapters of both languages.

Thankfully, prior linguistic studies provide exactly the information we need in the form of phylogeny trees. Relationships between languages are typically represented as tree or network diagrams. In the phylogenetic trees we will use, languages are grouped based on their similarities; an internal node may (but not necessarily) correspond to a hypothesized linguistic ancestor. While often a phylogenetic network is more appropriate than a tree (e.g. in cases of borrowing, or when two languages influence each other in a bidirectional manner), in this work we will focus on trees as a first step towards phylogeny-inspired adaptation.

\paragraph{Implementation} In a standard setting of adapting a language model from a source language to another target language, the typical approach~\cite[\textit{e.g.}][]{pfeiffer-etal-2020-mad} is to have source and target specific language adapters, trained separately on unlabeled monolingual text with the  masked language modeling (MLM) objective~\cite{devlin-etal-2019-bert}. Then, one can train a task adapter on source language task data, stacking it on top of the source language adapter. At evaluation time, the source language adapter is replaced with the target language one. 

As example, shown in Figure~\ref{fig:phylogeny}, consider three languages: Spanish, Catalan, and Asturian. To adapt a model for e.g. Named Entity Recognition (NER), the standard practice trains Spanish, Catalan, and Asturian language adapters separately: \texttt{L:Spanish}, \texttt{L:Catalan}, and \texttt{L:Asturian}. Using a language with labeled NER data (e.g. Spanish) then trains a task adapter \texttt{T:Spanish} using a stack of adapters \texttt{[L:Spanish, T:Spanish]}.  At inference time we can then use a stack with the appropriate language adapter to perform the task in that language e.g., stack \texttt{[L:Asturian, T:Spanish]}. 

Our approach follows the same principles, but adapters for multiple languages/genera/families are organized in a hierarchy following phylogenetic information and trained jointly. To continue with our running example, consider that all three languages belong to the Romance language group of the Indo-European family. We hence train five language type adapters jointly: \texttt{F:IndoEuro}, \texttt{G:Romance}, \texttt{L:Spanish}, \texttt{L:Catalan}, and \texttt{L:Asturian} which are stacked following the hierarchy depicted in Figure~\ref{fig:phylogeny}(c). So, examples from all IndoEuropean languages in our training mix are used to train the \texttt{F:IndoEuro} adapter, \texttt{G:Romance} is only trained on Romance languages data (if we have e.g. English or Danish in our mix, these data are directed through a \texttt{G:Germanic} adapter), and we also have language-dedicated adapters. We ensure that each training batch includes data from a single language; so, for an Asturian batch we train the following stack of adapters: \texttt{[F:IndoEuro, G:Romance, L:Asturian]}.  
At inference time, we also add the task adapter, trained as before on a language with labeled data, on top of our language-hierarchy adapters.


\section{Experimental Setup}
\paragraph{Tasks} We experiment on three NLP tasks: 
\begin{enumerate}
    \item Dependency Parsing (DEP),
    \item POS tagging (POS), and 
    \item Natural Language Inference (NLI). 
\end{enumerate}
For (1) and (2), we evaluate on 31 languages from Universal Dependencies v2.9 \cite{11234/1-4611}. For (3), we use 4 indigenous low-resource languages from AmericasNLI \cite{DBLP:journals/corr/abs-2104-08726}, an extension of XNLI \cite{conneau-etal-2018-xnli}. The choice of tasks and datasets is to ensure broad language coverage and especially to ensure we can study language families with only partial representation in the MLM pre-training stage. 

\paragraph{Language Families}
We study dependency parsing and POS-tagging on languages from the Germanic, Uralic and Tupian families.\footnote{To be accurate, the Germanic languages are a branch (genus) of the Indo-European family, not a distinct language family themselves.} For NLI, we work with languages from Uto-aztecan and Tupian families. See Appendix Table~\ref{tab:data_stat} for the complete list of languages we use to train family, group and language adapters.

\begin{table}[t]
\centering
\begin{tabular}{p{1.7cm}|p{3.2cm}|p{1.7cm}@{}}
\toprule
\textbf{Family} & \textbf{Genus} & \textbf{Tasks} \\ \midrule
Germanic & East Germanic, West Germanic & \small{POS, DEP} \\ \midrule
Uralic & Finnic, Hungarian, Permic, Mordvinic, Sami & \small{POS, DEP} \\ \midrule
Tupian & Tupari, Tupi-Guarani, Munduruku & \small{NLI, POS, DEP} \\ \midrule
Uto-Aztecan & Tepiman, Corachol, Yaqui, Aztecan, Tarahumaran & \small{NLI} \\ 
\bottomrule
\end{tabular}
\caption{Language families and genera we study.}
\label{tab:adapt_setting}
\vspace{-1em}
\end{table}

\paragraph{Pretraining Corpora}
For language adapter training we collect corpora from a variety of sources. See Appendix~\ref{app_sec:dataset} for the complete list of our data sources. As we experiment with a large number of low-resource and endangered languages, the number of sentences per language ranges from 3000 sentences to 1 million (i.e. the high resource ones). Following previous work, we experiment with up-sampling for the low-resource languages in our mix, to reduce data sparsity and to ensure they are adequately modeled.

\paragraph{Adapter Training}
For jointly training phylogeny-inspired adapters, we select training data from the language families/group presented in Table~\ref{tab:adapt_setting}.  Irrespective of task and setting, we train standard adapter architectures~\cite{ustun-etal-2020-udapter} leveraging the \texttt{AdapterHub.ml}~\cite{pfeiffer2020AdapterHub} framework. 

We train the task adapter by stacking it on top of the hierarchical language adapters. We follow the cross-lingual transfer setting of~\citet{pfeiffer-etal-2020-mad} where we select a high-resource language for task training: we use English for transfer for all families except Uralic, for which we switch to Estonian. In terms of base model choice, we use m\textsc{bert} for DEP, POS and \textsc{xlm-r} for NLI.\footnote{Results with both models for all tasks are available in Appendix: \ref{app_sec:result}.} For dependency parsing we train using the objective of~\citet{glavas-vulic-2021-supervised}, which is a modified variant of the standard deep biaffine attention dependency parser~\cite{DBLP:conf/iclr/DozatM17}. For all other tasks, we use simple classification heads as in previous literature.

\paragraph{Baselines and Model Variations}
We evaluate two common baselines for cross-lingual transfer:
\begin{enumerate}
    \item \texttt{[T]}: Using only the task adapter trained on some high-resource language; and
    \item \texttt{[LT]}: Using the stack of target language and task adapter. 
\end{enumerate}
We will denote our phylogeny inspired adapted models as \texttt{[FGLT]}: jointly trained \texttt{[Family, Group, Target Language]} stack and task adapter.
We also perform analyses and ablations without some parts of the task: for instance, \texttt{[FT]} and \texttt{[FGT]} denote stacks using only family (and genus) and task adapters without language-specific ones. 


\section{Results}


\begin{figure*}[t]
\small
    \centering
    \begin{tabular}{c}
        (a) Dependency Parsing \\
        \includegraphics[width=.9\textwidth]{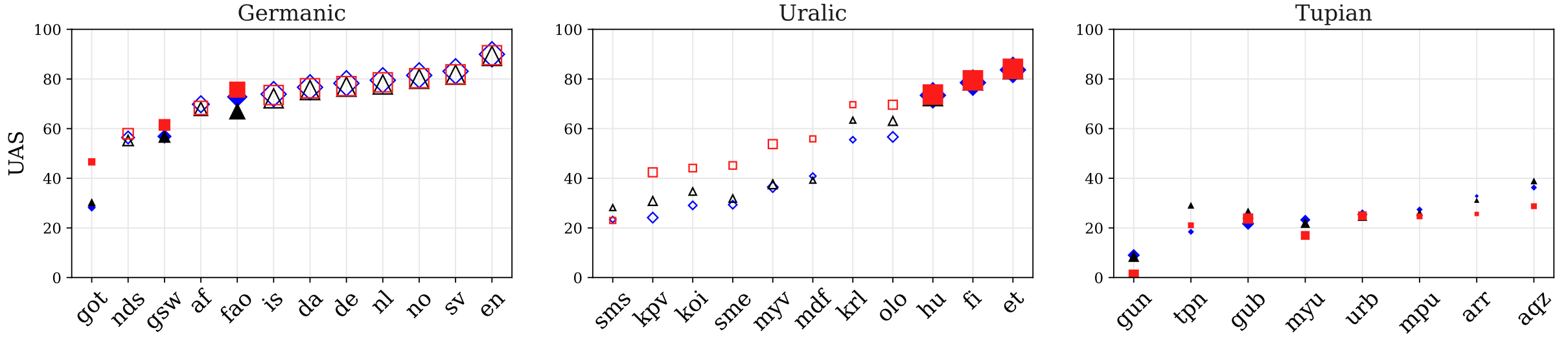}\\\\
        (b) POS tagging (POS)\\
        \includegraphics[width=.9\textwidth]{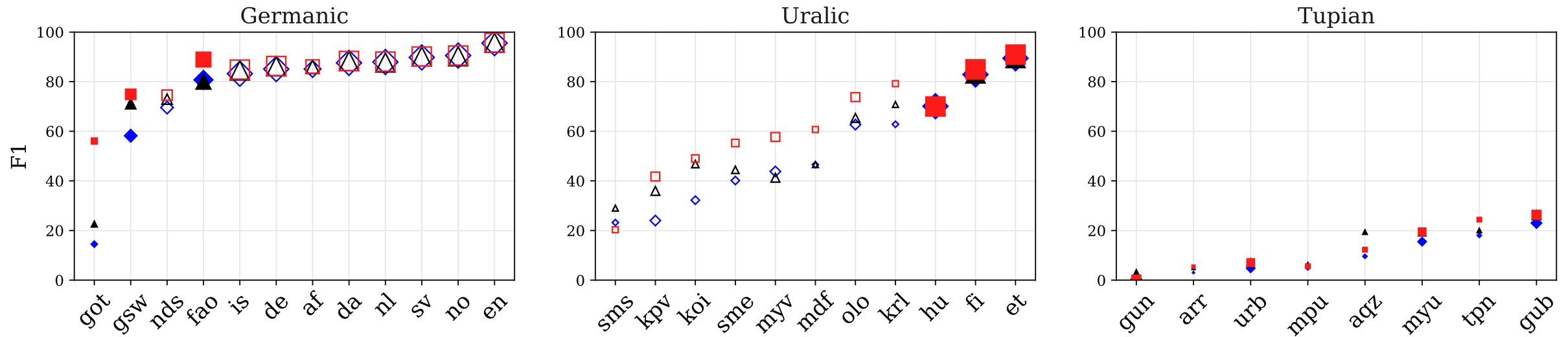}\\\\
        (c) Natural Language Inference (NLI) \\
        \includegraphics[width=.9\textwidth]{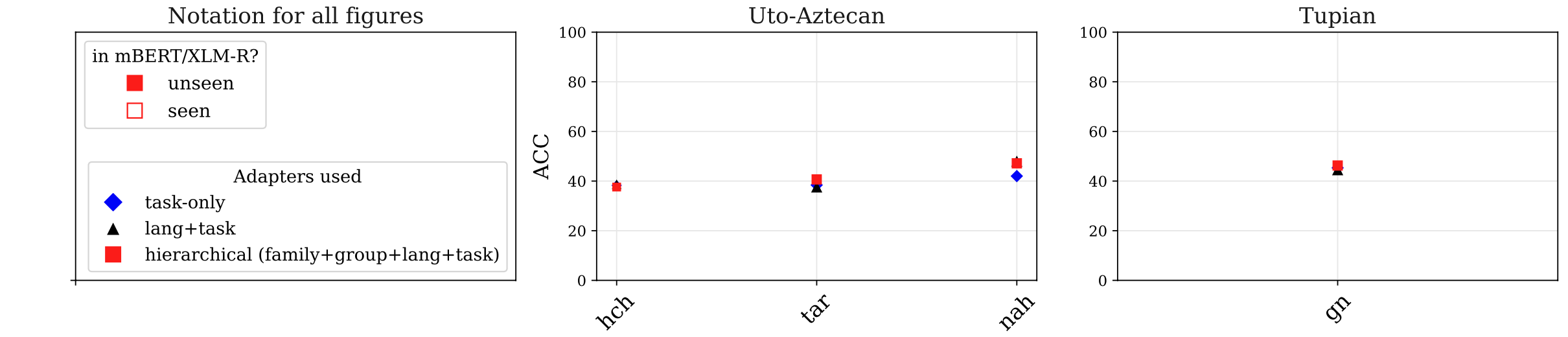}\\
    \end{tabular}
    \vspace{-1em}
    \caption{Visualizing three different task results across languages (marker size relative to MLM training data size).
In most cases, and especially in languages unseen during pre-training, our hierarchical phylogeny-inspired adapters
outperform the baselines.}
    \label{fig:result}
\end{figure*}
\begin{table*}[!t]
    \centering
    \small
    \begin{tabular}{@{ }r@{}l@{ }|c@{ \ }c@{ \ }c@{}c@{ \ }|c@{ \ }c@{ \ }c@{}c@{ \ }|c@{ \ }c@{ \ }}

\toprule
\ \ \textbf{Task (metric):}&&\multicolumn{3}{c}{\textbf{Dep-Parsing (UAS)}}& &\multicolumn{3}{c}{\textbf{POS-Tagging (F1-score)}}&&\multicolumn{2}{c}{\textbf{NLI (Acc.)}} \\
\midrule
\ \ Language &-Family&  Germanic & Uralic & Tupian & & Germanic & Uralic & Tupian& & Uto-Aztecan & Tupian \\
\ \ \begin{tabular}{@{}c@{}}Language \\ (Unseen,\end{tabular} &\begin{tabular}{@{}c@{}}-Count \\ Total)\end{tabular} &  (3,12) & (8,11) & (8,8) & & (3,12) & (8,11) & (8,8) & & (3,3) &
(1,1) \\
\midrule 
\multicolumn{3}{l}{\textbf{Baselines}}\\
\ \ \textsc{Base-LM} & \multirow{1}{*}{+ \texttt{[T]} }  & 52.5 (70.6) &  36.9 (48.3) & \textbf{24.1} & &  51.1 (77.3) & 41.9 (52.5) & 9.9 & &  39.6 & 45.3  \\
\ \ \textsc{Base-LM} & \multirow{1}{*}{+ \texttt{[LT]} }  & 50.8 (69.2) &  41.1 (51.4) & 19.0 & &  57.9 (79.6) & 47.5 (56.7) & \textbf{13.2} & &  41.3 & 44.4  \\
\multicolumn{3}{l}{\textbf{Phylogenically inspired}}&\\
\ \ \textsc{Base-LM} & \multirow{1}{*}{+ \texttt{[FGLT]} }  & \textbf{60.1} (\textbf{72.3}) &  \textbf{50.5}  (\textbf{58.3}) & 23.8 & &  \textbf{73.3} (\textbf{83.7}) & \textbf{54.7} (\textbf{62.2}) & 12.6 & &  \textbf{41.8} & \textbf{46.3} \\
\bottomrule
\end{tabular}

    \caption{Average results per language family across different tasks. We report averages both for languages unseen during pretraining, and for all languages in the mix (the latter in parentheses). Base language model (\textsc{Base-LM}) is mBERT for Dep-Parsing, POS-Tagging and XLM-R for NLI. We use the following language for task adapter training: English for Germanic, Tupian and Uto-Aztecan and Estonian for Uralic.}
    \label{tab:avg}
\end{table*}

\paragraph{General Observations}
We present our experimental results covering all three tasks in Table~\ref{tab:avg}, showing average performance for the baselines and our proposed method. We further split the results for languages seen and not seen by m\textsc{bert} during pretraining. Compared to the \texttt{[T]} and \texttt{[LT]} baselines, we observe substantial performance improvements in 10 out of 12 task-family specific settings using \texttt{[FGLT]}. A visualization of all three task results with a breakdown per language is also available in Figure~\ref{fig:result}.

Looking at Figure~\ref{fig:result}, it is quite apparent how phylogeny inspired adaptation uplifts the performance of low-resource languages, especially the ones unseen during pretraining. For example, we evaluate dependency parsing on 3 such Germanic languages (Faroese, Gothic and Swiss German). All 3 languages benefit from the proposed adaptation approach with maximum 16.46\% improvement over the best performing baseline for Gothic (see Table~\ref{tab:test_dep}). 

This positive drift of performance becomes more obvious for Uralic languages. Here, 8 out of 11 languages are extremely low-resource ones and unseen during pretraining. We obtain improvements over baseline in 7 out of these 8. We further observe similar trends in POS-Tagging for both Germanic and Uralic languages irrespective of the choice of base language model (see Appendix Tables~\ref{tab:test_dep}---\ref{tab:test_pos_xlmr}). 

The other language families we focus on are Tupian, Uto-Aztecan, comprised of indigenous and very low-resource languages~\cite{DBLP:journals/corr/abs-2104-08726}. In case of Tupian languages on DEP-Parsing and POS-Tagging, we observe model adaptation does not result in improvement over baselines on mBERT. However, when we use XLM-R with model adaptation, average performance improves all around for these two tasks. In addition, for NLI, which is a task requiring higher semantic capabilities, we conduct experiments on four languages from Uto-Aztecan and Tupian families. As before, the combination of XLM-R with phylogenetic adaptation outperforms all other settings. 


Among the baselines, the task-adapter-only baseline \texttt{[T]} performs better in Germanic and Tupian DEP-Parsing compared to the \texttt{[LT]} baseline. This points out the known problems with negative interference~ \cite[\textit{inter alia}]{8953565,wang-etal-2020-negative}. On the contrary, token classification tasks like POS-Tagging gets significant benefits from using the \texttt{[LT]} baseline. Compared to these, \texttt{[FGLT]} leads to consistent performance improvements. Even though our method does not uplift the result for Tupian DEP-Parsing and POS-Tagging, it is worth noting that it does not hurt either, unlike  e.g. \texttt{[T]} which hurts in DEP-Parsing (-0.3 points compared to -5.1 points). Last, outperforming the average baseline of four indigenous American languages~\cite{DBLP:journals/corr/abs-2104-08726}, points out the effective adaptation capabilities of phylogeny-based adaptation. See Appendix~\ref{app_sec:result} for detailed language specific results.

\paragraph{True Zero-Shot Adaptation}
\begin{table}[!t]
    \centering
    \small
    \begin{tabular}{@{ }l@{}l@{ }|c@{ \ }c@{ \ }c@{ \ }c@{ \ }|c@{}}

\toprule
\multicolumn{7}{c}{\textbf{Uralic (language adpater: est)}} \\
 Model & Training & sms &    mdf &    krl &&    avg \\
\midrule
\multicolumn{7}{l}{\textbf{Baselines}}\\
 \ \ \textsc{mbert} & \multirow{1}{*}{+ \texttt{[T]} (est)}  &  23.37 &  40.89 &  55.53 &&  39.93 \\
 \ \ \textsc{mbert} & \multirow{1}{*}{+ \texttt{[LT]} (est)}  &  23.82 &  41.08 &  53.68 &&  39.53 \\
  \multicolumn{7}{l}{\textbf{Phylogenically inspired}}\\
\ \ \textsc{mbert} & \multirow{1}{*}{+ \texttt{[FGLT]} (est)}  &   23.74 &  \textbf{42.01} &  53.98 &&  39.91 \\
 \multicolumn{7}{l}{\textbf{Ablations}}\\
  \ \ \textsc{mbert} & \multirow{1}{*}{+ \texttt{[FT]} (est)}  &  \textbf{25.81} &  39.37 &  57.18 &&  40.78 \\
 \ \ \textsc{mbert} & \multirow{1}{*}{+ \texttt{[FGT]} (est)}  &  24.48 &  41.35 &  \textbf{58.99} &&  \textbf{41.60} \\

\toprule
\multicolumn{7}{c}{\textbf{Tupian (language adpater: gub)}} \\
 Model & Training & aqz &    mpu &    tpn &    urb &    avg \\
\midrule
\multicolumn{7}{l}{\textbf{Baselines}}\\
 \ \ \textsc{mbert} & \multirow{1}{*}{+ \texttt{[T]} (eng)}  &  27.50 &  \textbf{23.97} &  22.37 &  24.59 &  24.61 \\
 \ \ \textsc{mbert} & \multirow{1}{*}{+ \texttt{[LT]} (eng)}  &  22.50 &  17.81 &   9.21 &  25.41 &  18.73 \\
 \multicolumn{7}{l}{\textbf{Phylogenically inspired}}\\
\ \ \textsc{mbert} & \multirow{1}{*}{+ \texttt{[FGLT]} (eng)}  &   \textbf{27.50} &  19.86 &  \textbf{31.58} &  \textbf{26.78} &  \textbf{26.43} \\
 \multicolumn{7}{l}{\textbf{Ablations}}\\
 \ \ \textsc{mbert} & \multirow{1}{*}{+ \texttt{[FT]} (eng)}  &  21.25 &  17.81 &  14.47 &  17.76 &  17.82 \\
 \ \ \textsc{mbert} & \multirow{1}{*}{+ \texttt{[FGT]} (eng)}  &  22.50 &  17.12 &  19.74 &  22.13 &  20.37 \\

\bottomrule
     \end{tabular}
    \caption{Dependency parsing with extremely low-resource languages in the absence of language specific adapters (true zero-resource scenario).}
    \label{tab:ex_lr}
    \vspace{-1em}
\end{table}


For a large number of extremely low-resource languages not seen during the pre-training of current language models, there may be no easily obtainable textual data to even perform MLM training to train a language-specific adapter. We explore such a scenario and investigate whether the language-family adaptors can be used instead of language-specific ones.

We simulate this scenario in two settings. First for 3 Uralic languages: Skolt Sami \texttt{(sms)}, Moksha \texttt{(mdf)} and Karelian \texttt{(krl)}. We discard their data from the training set and train other adapters jointly as before. During evaluation, we just use a high-resource language adapter (\texttt{L:Estonian}) instead of the missing language adapters. In addition, we explore this scenario in 4 Tupian languages: Akuntsu \texttt{(aqz)}, Makuráp \texttt{(mpu)}, Tupinambá \texttt{(tpn)} and Kaapor \texttt{(urb)} where we actually do not have any available training data (except \texttt{(urb)}. So we replace the language adapter with a higher-resource one \texttt{(L:Guajajára)}. 

Results are presented in Table~\ref{tab:ex_lr}. Looking at the rows with phylogenically inspired adaptation \texttt{[FGLT]}, we see 1.82\% improvement on average for Tupian languages over the best performing baseline (\texttt{[T]}). Except Makuráp \texttt{(mpu)}, all other 3 Tupian languages benefit from using our family adapters. Perhaps the most important result is the one on Tupinambá \texttt{(tpn)} which gets drastically impacted when using only baseline language adapter \texttt{[LT]}(-13.16\% from \texttt{[T]}) but performs much better with \texttt{[FGLT]}(+9.21\% over \texttt{[T]}).  

\begin{table*}[t]
\small
    \centering
    \begin{tabular}{@{ }l@{}l@{ }|c@{ \ }c@{ \ }c@{ \ }|c@{ \ }c@{ \ }c@{ \ }c@{ \ }c@{ \ }c@{ \ }c@{ \ }c@{ \ }|c@{}}

\toprule
\multicolumn{14}{c}{\textbf{Uralic (DEP-Parsing)}} \\ \midrule
&&\multicolumn{3}{c|}{\textbf{\textsc{mbert-seen}}} & \multicolumn{8}{c|}{\textbf{\textsc{mbert-unseen}}}&\\
Model&Training& est & fin & hun & koi & kpv & krl & mdf & myv & olo & sme & sms & avg \\
\midrule 
\multicolumn{4}{l}{\textbf{Adapter Parameter count: constrained (885312)}} \\

 \ \ \textsc{mbert} & \multirow{1}{*}{+ \texttt{[LT]} (est)}  &  84.05 &  79.08 &  73.00 &    32.30 &    26.85 &   53.52 &   37.52 &    35.08 &    54.30 &    26.23 &   25.89 &  47.98 \\
 \ \ \textsc{mbert} & \multirow{1}{*}{+ \texttt{[LLLT]} (est)}  &  86.01 &  79.51 &  74.47 &    32.30 &    27.71 &   49.23 &   37.39 &    33.34 &    51.21 &    25.73 &   20.56 & 47.04  \\
  \ \ \textsc{mbert} & \multirow{1}{*}{+ \texttt{[FGLT]} (est)}  &  83.23 &  78.48 &  72.63 &    37.43 &    32.21 &   64.06 &   44.12 &    39.79 &    64.78 &    30.75 &   24.26 &  51.98 \\
 \multicolumn{6}{l}{\textbf{Adapter Parameter count: Upscaled (2655936 or, 3$\times$885312)}} \\ 
 \ \ \textsc{mbert} & \multirow{1}{*}{+ \texttt{[FGLT]} (est)}  &  84.20 &  79.59 &  73.10 &    38.14 &    35.55 &   65.77 &   44.52 &    42.77 &    67.94 &    31.62 &   22.78 &  53.27 \\
\bottomrule
     \end{tabular}
    \caption{Effect of parameter reduction in dependency parsing (Metric: UAS) on Uralic languages.}
    \label{tab:param_c}
    \vspace{-1em}
\end{table*}

For Uralic languages, even our model ablations (shown in Table~\ref{tab:ex_lr}) perform better than the baselines: these are \texttt{[FT]} and \texttt{[FGT]} where we get rid of the language adapter part and just draw inference from family and genre adapters. Specifically, \texttt{[FGT]} shows consistent improvement for all 3 Uralic languages, even though the model never observed the target language texts during neither base model pretraining nor adapter training. 



\section{Further Discussion}

We perform additional ablation studies where we show that our proposed approach provides sustainable performance in constrained settings with reduced parameter counts. In addition, we explore data up-sampling for low-resource languages in language families with large data imbalances across the language members. This simple approach points towards the further improvement scope with limited data availability. Detailed analysis of both these experiments are presented below.

\paragraph{Parameter Reduction}
Stacking multiple adapters instead of a single language adapter comes with extra parameter cost.\footnote{We note, though, that this additional cost is still a very small fraction of the overall model's parameter count.} To assess whether we can integrate phylogenetic information while keeping the adapter parameter counts limited, we perform parameter reduction using a constant factor. For example, consider a single language adapter \texttt{[L]} which has down/upword projections with \texttt{L:Proj}$\times$\texttt{Layer} parameters leading to a parameter count of \texttt{2}$\times$\texttt{48}$\times$\texttt{768}. Instead we can use a dimension reduced by a factor of 3 and add two extra adapters (\texttt{[FGL]}) without increasing the parameter count \texttt{2}$\times$(\texttt{F:Proj+G:Proj+L:Proj})$\times$\texttt{FGL:Output}; to be accurate: \texttt{2}$\times$(\texttt{16+16+16})$\times$\texttt{768}. Contrast these with our solution without this constant factor parameter reduction, which will add \texttt{2}$\times$(\texttt{48+48+48})$\times$\texttt{768} parameters to be learned.


The results, tested on Uralic languages for the dependency parsing task, are reported in Table~\ref{tab:param_c}. Importantly, we observe consistent performance improvement in \texttt{[FGLT]} over baseline \texttt{[LT]} irrespective of the parameter count. Among these two selections, the \texttt{[FGLT]} one with constrained parameter count (885312) comes with a 1.29\% performance trade off which still outperforms the baseline by 4 points on average. Further looking into each individual language result, we find an interesting trend in Skolt Sami (\texttt{sme}). This is the only language where performance drops in constrained \texttt{[FGLT]} compared to the baseline which then drops further when we move to the upscaled \texttt{[FGLT]}. Likewise, we observe performance improvement in any language using sustained model elevates further in upscaled model. 

\paragraph{Deep vs Wide Adapters} Our \texttt{FGLT} setting makes two important changes to the baseline \texttt{LT} one. First, it stacks 3 language-related adapters as opposed to a single one. Second, it shares some of these adapters between languages. An important question is whether the performance improvements are due to stacking (making the model \textit{deeper}) or due to the parameter sharing between languages. 
To answer this question, we perform another ablation where we replace the \texttt{2}$\times$(\texttt{F:Proj+G:Proj+L:Proj})$\times$\texttt{FGL:Output} setting with \texttt{2}$\times$(\texttt{L:Proj+L:Proj+L:Proj})$\times$\texttt{LLL:Output}. 
Essentially, we create a stack of 3 language-specific adapters.

We will first contrast the baseline \texttt{[LT]} (which has a single \textit{wide adapter}) to this deeper version \texttt{[LLLT]}. We keep the parameter count equal between the two using the same parameter reduction as in the previous paragraph.
We find that the \texttt{[LLLT]} setting does indeed improve performance, but only for high-resource languages, even exceeding the upscaled phylogenetic setting \texttt{[FGLT]} (see Table~\ref{tab:param_c}). For 7 out of 8 low-resource languages unseen by mBERT, however, the performance degrades in \texttt{[LLLT]} compared to \texttt{[LT]}. 
Hence, we conclude that deeper stacks of adapters are better than a single wide adapter, but without the adapter parameter sharing this only benefits high-resource languages.


We want to further focus on this second point about parameter sharing:
in Table~\ref{tab:param_c}, compare rows \texttt{[LLLT]} and \texttt{[FGLT]} under the reduced parameter count. For \textit{all} unseen languages, \texttt{[FGLT]} yields significant improvements, leading to almost 5 UAS points higher on average.

\paragraph{Effect of Upsampling}
\begin{table*}[t]
\small
    \centering
    \begin{tabular}{@{ }l@{}l@{ }|c@{ \ }c@{ \ }c@{ \ }c@{ \ }c@{ \ }c@{ \ }c@{ \ }c@{ \ }c@{ \ }c@{ \ }c@{ \ }|c@{}}

\toprule
 Model & Training & sme &  koi &  fin* &  myv &  olo &  mdf &  hun* &  sms &  kpv &  est* &  krl &    avg \\
\midrule 
\multicolumn{2}{l}{\textbf{Original datasize}:} &10k & 10k & 1M & 29k & 19k & 5k & 1M & 3k & 13k & 1M & 5k\\
 \ \ \textsc{mbert} & \multirow{1}{*}{+ \texttt{[FGLT]} (et)}  &  31.62 &    38.14 &  79.59 &    42.77 &    67.94 &   44.52 &  73.10 &   22.78 &    35.55 &  84.20 &   65.77 &  53.27 \\
\midrule
\multicolumn{2}{l}{\textbf{Upsampled}:} &100k & 60k & 1M & 87k & 116k & 28k & 1M & 29k & 40k & 1M & 36k\\
 \ \ \textsc{mbert} & \multirow{1}{*}{+ \texttt{[FGLT]} (et)}  &  45.16 &    44.10 &  79.45 &    53.77 &    69.62 &   55.88 &  73.73 &   23.00 &    42.40 &  84.10 &   69.65 &  58.26 \\

\bottomrule
     \end{tabular}
    \caption{Dependency parsing result (UAS) upsampling datasize (* columns are the high-resourced ones and not up-sampled, the presented datasize is approximate sentence count per language)}
    \label{tab:upsample}
\end{table*}


\begin{table}[!t]
    \centering
    \small
    \begin{tabular}{@{ }l@{}l@{ }|c@{ \ }c@{ \ }c@{ \ }|c@{}}

\toprule
 Model & Training & fao &    kpv &    urb &    avg \\
 \midrule
 \multicolumn{6}{c}{\textbf{DEP }  (task adpater: eng)} \\
\midrule
\multicolumn{6}{l}{\textbf{Baselines}}\\
 \ \ \textsc{mbert} & \multirow{1}{*}{+ \texttt{[T]} }  &  72.80 &  24.15 &  24.59 &  40.51 \\
 \ \ \textsc{mbert} & \multirow{1}{*}{+ \texttt{[LT]}}  &  66.93 & 30.87 &  25.41 &  41.07 \\
  \multicolumn{6}{l}{\textbf{Phylogenically inspired} }\\
\ \ \textsc{mbert} & \multirow{1}{*}{+ \texttt{[FGLT]} }  &   \textbf{75.70} &  \textbf{42.40} &  \textbf{26.78} &  \textbf{48.29} \\
 \multicolumn{6}{l}{\textbf{Random Tree}}\\
  \ \ \textsc{mbert} & \multirow{1}{*}{+ \texttt{[FGLT]}}  &  66.19 &  28.53 & 24.04 &  39.59 \\

\midrule
\multicolumn{6}{c}{\textbf{POS } (task adpater: eng)} \\
\midrule
\multicolumn{6}{l}{\textbf{Baselines}}\\
 \ \ \textsc{mbert} & \multirow{1}{*}{+ \texttt{[T]} }  &  80.70 &  24.02 &  4.79 &  36.50 \\
 \ \ \textsc{mbert} & \multirow{1}{*}{+ \texttt{[LT]}}  &  79.93 &  35.96 &  7.13 &  41.01 \\
  \multicolumn{6}{l}{\textbf{Phylogenically inspired}}\\
\ \ \textsc{mbert} & \multirow{1}{*}{+ \texttt{[FGLT]} }  &   \textbf{88.88} &  \textbf{41.74} &  7.10 &  \textbf{45.91} \\
 \multicolumn{6}{l}{\textbf{Random Tree}}\\
  \ \ \textsc{mbert} & \multirow{1}{*}{+ \texttt{[FGLT]} }  &  86.66 &  35.96 &  \textbf{13.66} &  45.43 \\

\bottomrule
     \end{tabular}
    \caption{Adapters arranged following a phylogenetically-inspired tree perform significantly better than ones following random counterfactual tree. Parameter sharing between similar languages leads to significantly better results for the unseen languages in both tasks.}
    \label{tab:rand_res}
    \vspace{-1em}
\end{table}
For most of the Uralic, Germanic and all of the Tupian and Uto-Aztecan low-resource languages, we had very little amount of training data available. As a result, this limited data availability creates within-family data imbalance, especially for Germanic and Uralic languages. To address this issue, we perform a simple data upsampling on all low resource languages from these two families. Here, the upsampling factor is inversely proportional  to the per-language token count. A language with very low word count is sampled in large numbers compared to the ones with higher word count.  

We use the upsampled dataset for all the dependency parsing and POS tagging experiments we perform on these two language families (Appendix Table \ref{tab:avg}, \ref{tab:test_dep}, \ref{tab:test_dep_xlmr}, \ref{tab:test_pos}, \ref{tab:test_pos_xlmr}). The positive upsampling effect is obvious when we compare the dependency parsing results on Uralic upsampled dataset with the one with original datasize in Table~\ref{tab:upsample}. Note that we do not upsample the 3 high resource ones: Estonian (\texttt{et}), Finnish (\texttt{fi}), and Hungarian (\texttt{hu}) and experiment on the other languages, where we can make a number of interesting observations. 

First, though the original sentence count is same (10k) for North Sami (\texttt{sme}) and Komi Permyak (\texttt{koi}) the upsampled size is different for these two languages: 100k and 60k respectively. The reason behind this difference is, we perform word-count based upsampling and the average sentence length turns out to be less for \texttt{koi} thus assigned with a low sampling factor. Hence, the one with higher upsampled sentence count (\texttt{sme}) results in large performance improvement of 13.54 points, while it was the one with second lowest score in the non-upsampled setting. Secondly, we observe performance improvements for all low-resource languages. It would be interesting to explore the resource dependent performance variation that could be attributed to data sampling choices. For now, we keep this open for future studies. 

On the other hand, we cannot clearly claim that extremely low-resource languages always benefit from upsampling. For example, Skolt Sami (\texttt{sms}) is the one with lowest data availability (3k) and lowest original score (22.78). Upsampling more than 9x times results in only 0.22\% improvement. We suspect that data quality might play an important role here, considering  that we had to scrape the few data available online for \texttt{sms}~\cite{wanca}, whereas the corpus we use for \texttt{sme} was collected by~\citet{goldhahn-etal-2012-building} following standard approaches and with NLP applications in mind.

\paragraph{Random vs Phylogenetic Tree}
One key hypothesis of ours is that language family tree information is beneficial for modeling low-resource languages. 

To further solidify this claim, we compare adapters based on a linguistically-informed tree (like the one we have been using in all previous experiments) to adapters based on a counter-factual (hypothetical) language tree.
We construct a random language family hierarchy and train the adapter stacks jointly like before instead of using the phylogenetically informed ones. 
We make the random tree structure typologically diverse while keeping one low-resource language from either Germanic (Faroese), Uralic (Komi Zyrian) or Tupian (Kaapor) present in each newly defined genus (see Table~\ref{tab:rand_tree} in Appendix \ref{app_sec:rand} for the random family tree structure). In Table~\ref{tab:rand_res}, we report results in Dependency parsing and POS tagging tasks for these 3 languages under each of these settings. The results for dependency parsing are to a large extent conclusive: the adapters following the random tree perform worse than the baselines, while the phylogenetically-inspired ones are significantly better. The random-tree adapters do indeed outperform the baselines for POS tagging, but again for 2 of the 3 low-resource languages fall short compared to the phylogenetically-inspired ones. Curiously, for Kaapor, this random-tree model outperforms all other models, but all of them are still extremely bad (with only an accuracy of 13\% in the best case); nevertheless, we will further investigate this result in future work.

\paragraph{Indo-European Family Tree}
Going beyond our original setup, we conduct one additional experiment where we do joint-training on the whole Indo-European language family as shown in Figure~\ref{fig:phylogeny}. The only difference is that essentially, by adding a \textit{root} adapter \texttt{R} we have a stack of four jointly trained adapters  \texttt{[RFGL] (R:IndoEuro)} instead of just three (i.e. \texttt{[FGL]}). Interestingly, the performance on the dependency parsing tasks gets negatively impacted for almost all languages (see Table~\ref{tab:ie_mbert_dep}). 
We hypothesize that this is due to the inherent diversity of the Indo-European family. Despite sharing a common ancestor (Proto-Indo-European), the IE family groups that we work with here (Germanic, Romance, Slavic, Celtic, Greek, Indo-Aryan) are too typologically different from each other, and forcing them to share a common root negates the gains of the group-specific adapters. We plan to investigate this further in future work.


\section{Related Work}
\label{app:related}
Continuous effort is being put to improve cross-lingual transfer across languages as well as making language models capable enough to go beyond high resource domains. Recently, \citet{lexicon_wang}, proposed an approach to combine lexicons with monolingual/parallel data for pretraining. It expands the modeling capability to thousands more languages largely including under-represented languages with limited to zero corpus availability. It is now proven that, pretraining on closely related languages yields better result for zero-shot transfer \cite{pires-etal-2019-multilingual} and continued pretraining on a larger number of languages leads to further improvement \cite{fujinuma-etal-2022-match}. However, training on some specific languages can still hurt the performance of other languages \cite{conneau-etal-2020-unsupervised}. As a result, it is crucial to prevent negative inference while keeping the performance equitable and robust across languages~\cite{8953565, wang-etal-2020-negative}.

To make the performance robust across languages, it is important to identify how much linguistic information is currently in place inside these big multilingual models. Recent studies have done investigation on this hypothesis by probing language models for linguistic
typology \cite{10.1162/coli_a_00444, https://doi.org/10.48550/arxiv.2205.02023} as well as phylogheny \cite{rama-etal-2020-probing}. These studies have measured phylogenetic distance and typological similarity across languages so that we can make informed cross-lingual transfer. In line with these findings, \cite{zhao-etal-2021-inducing} has done experiments to remove the language specific information by stackable vector operations which further improve the cross-lingual representation. One recent study \cite{https://doi.org/10.48550/arxiv.2205.12672} dives further into identifying language-neutral and language-specific subspace inside the representation space of multilingual models and now it is proven that the shared representation space is the one helping to perform effective cross-lingual transfer. 

As opposed to the standard fine-tuning of large-scale language models, a more focused trend is to perform efficient parameter selection thus reducing the overall computation cost and carbon footprints~\cite{https://doi.org/10.48550/arxiv.1902.00751}. Adapters are such highly customized light-weight neural network layers on top of base models. Because of this higher flexibility, there are studies already in place looking into the adapter-level optimization according to the nature of data and network layers~\cite{adaptable_adapters}. In addition, using language specific units in a modular fashion in the pre-training stage was shown to be beneficial in recent work \cite{pfeiffer2022xmod}.

\section{Limitations and Future Work}
While we already incorporated task evaluation on a diverse set of language families ranging from extremely low resourced Uralic ones to indigenous AmericasNLI~\cite{DBLP:journals/corr/abs-2104-08726} languages, our experiments are still limited in terms of typological diversity. In future, we want to further extend the typological diversity of languages we use. At the same time, we would like to democratize the full force of language genetical properties in  steps beyond just finetuning thus making the resource scarce languages more accessible. 

\section{Conclusion}
In this work, we present an adapter-based approach to leverage language phylogenetic information for better cross-lingual adaptation. Our experiments on a diverse set of tasks and languages show significant performance improvements over commonly-used strong baselines. Even better, we show that under the exact same adapter parameter count settings, using smaller adapters but forcing adapter sharing between genetically related languages improves performance on true zero-resource scenarios. These improvements are particularly stark for languages unseen in the pre-training stage of large multilingual language models, providing a direct path towards better adaptation and language coverage for language technologies. 



\section*{Acknowledgements}

This work is generously supported by NSF Award IIS-2125466 and by a Google Award for Inclusion Research.

\bibliography{anthology,acl}
\bibliographystyle{acl_natbib}
\clearpage
\newpage

\appendix

\section{Dataset}
Detailed data source with statistics are presented in table \ref{tab:data_stat}.
\label{app_sec:dataset}
\begin{table*}[!t]
    \centering
    \small
    \begin{tabular}{llllll}
\toprule
      Family &        Genus &              Language &                ISO 639-3 &  Size &  Source \\
\midrule
   \multirow{12}{*}{Germanic} &         North &              Danish &              dan &     1M &       OSCAR~\cite{OrtizSuarezSagotRomary2019} \\
     &         North &           Faroese &           fao &     300K &       \cite{goldhahn-etal-2012-building} \\
     &         North &              Icelandic &              isl &     1M &       OSCAR~\cite{OrtizSuarezSagotRomary2019} \\
     &         North &              Norwegian &              nor &     1M &       OSCAR~\cite{OrtizSuarezSagotRomary2019} \\
     &         North &              Swedish &              swe &     1M &       OSCAR~\cite{OrtizSuarezSagotRomary2019} \\
     &          West &                 Afrikaans &                 afr &     120K &       OSCAR~\cite{OrtizSuarezSagotRomary2019}\\
     &          West &              German &              deu &     1M &       OSCAR~\cite{OrtizSuarezSagotRomary2019} \\
     &          West &              English &              eng &     1M &       OSCAR~\cite{OrtizSuarezSagotRomary2019} \\
     &          West &             Gothic &             got &     4.4K &       Bible~\cite{wulfila} \\
     &          West &          Low Saxon &          nds &     95.5K &       \cite{siewert-etal-2020-lsdc} \\
     &          West &              Dutch &              nld &     1M &       OSCAR~\cite{OrtizSuarezSagotRomary2019} \\
     &          West &  Swiss German &  gsw &     100K &       \cite{goldhahn-etal-2012-building} \\
     \midrule
      \multirow{6}{*}{Tupian} &     Munduruku &                Munduruku &                myu &     8.7K &       Bible~\cite{splash} \\
       &  Tupi Guaraní &                Guaraní &                 grn &     26K & \cite{chiruzzo-etal-2020-development} \\
       &  Tupi Guaraní &                Simba Guaraní &                gnw &     6.7K &       Bible~\cite{splash}\\
       &  Tupi Guaraní &                Guajajára &                gub &     33.9K &     Bible~\cite{splash} \\
       &  Tupi Guaraní &                Mbya Guaraní&                gun &     50.5K &       Bible~\cite{splash} \\
       &  Tupi Guaraní &                Kaapor &                urb &     9.3K &       Bible~\cite{splash} \\
       &  Tupari &                Akuntsu &                aqz &     - &       - \\
       &  Tupari &                Makuráp &                mpu &     - &       - \\
       &  Tupi-Guarani &                Tupinambá &                tpn &     - &       - \\
       \midrule
      \multirow{11}{*}{Uralic} &        Finnic &              Estonian &              est &     1M &       OSCAR~\cite{OrtizSuarezSagotRomary2019} \\
       &        Finnic &              Finnish &              fin &     1M &       OSCAR~\cite{OrtizSuarezSagotRomary2019} \\
       &        Finnic &             Karelian&             krl &     5K &       Bible~\cite{krl} \\
       &        Finnic &            Livvi &            olo &     19K&  \cite{vepkar} \\
       &     Hungarian &              Hungarian&              hun &     1M &       OSCAR~\cite{OrtizSuarezSagotRomary2019} \\
       &     Mordvinic &             Moksha&             mdf &     5K &       Bible~\cite{krl} \\
       &     Mordvinic &            Erzya &            myv &     29K&       \cite{myv} \\
       &        Permic &            Komi Permyak &     koi &     10K&       \cite{goldhahn-etal-2012-building} \\
       &        Permic &            Komi Zyrian &      kpv &     13K&   \cite{kpv_lit} \\
       &          Sami &            North Sami &       sme &     10K&   \cite{goldhahn-etal-2012-building} \\
       &          Sami &             Skolt Sami &      sms &     3K&\cite{wanca} \\
       \midrule
 \multirow{8}{*}{Uto-Aztecan} &       Aztecan &                Nahuatl &                nah &     16K &  \cite{gutierrez-vasques-etal-2016-axolotl} \\
  &      Corachol &                Cora &                crn &     10.1K &       Bible~\cite{splash} \\
  &      Corachol &                Huichol &                hch &     8.9K &\cite{hch} \\
  &   Tarahumaran &                Rarámuri &                tar &     14.7K & \cite{brambila} \\
  &       Tepiman &                Northern Tepehuan &                ntp &     6.5K &       Bible~\cite{splash} \\
  &       Tepiman &                O’odham &                ood &     6.5K &       Bible~\cite{splash} \\
  &       Tepiman &                Southern Tepehuan &                stp &     7K &       Bible~\cite{splash} \\
  &         Yaqui &                Mayo &                mfy &     7K &       Bible~\cite{splash} \\
  &         Yaqui &                Yaqui &                yaq &     6.5K &       Bible~\cite{splash} \\
\bottomrule
\end{tabular}
    \caption{Dataset statistics and sources of the language datasets we work with.}
    \label{tab:data_stat}
\end{table*}

\section{Language Specific Task Results}
\label{app_sec:result}
Detailed language specific task results are presented in table \ref{tab:test_dep}, \ref{tab:test_dep_xlmr}, \ref{tab:test_pos}, \ref{tab:test_pos_xlmr}, \ref{tab:test_nli_mbert} and \ref{tab:test_nli}.
\paragraph{Dependency Parsing} 
For dependency parsing, we perform experiments on Germanic, Uralic and Tupian languages. We observe, phylogeny based joint training performs better for 10 out of 11 Germanic and Uralic languages unseen by mbert. In addition all of the Tupian ones are unseen by mbert and joint training performs better than the language based adapter baseline \texttt{[LT]}. Similar trend is visible in case of Germanic high resource languages where using the language based adapter baseline \texttt{[LT]} hurts the overall performance. Though, joint training does not cross the performance threshold of just using the task adapter baseline \texttt{[T]} in case of majority high resource ones, it doesn't do negative interference like language adapter based baseline either. At the same time, the performance improvement for unseen low resource languages are significant while using joint training. Thus phylogeny based joint training keeps a performance balance across languages with diverse data availability.
\begin{table*}[!t]
    \centering
    \small
    \begin{tabular}{@{ }l@{}l@{ }|c@{ \ }c@{ \ }c@{ \ }c@{ \ }c@{ \ }c@{ \ }c@{ \ }c@{ \ }c@{ \ }|c@{ \ }c@{ \ }c@{ \ }|c@{}}

\toprule
\multicolumn{15}{c}{\textbf{Germanic}} \\
\midrule
&&\multicolumn{9}{c}{\textbf{\textsc{mbert-seen}}} & \multicolumn{3}{c}{\textbf{\textsc{mbert-unseen}}}&\\
Model&Training& afr & dan & deu & eng & isl & nds & nld & nor & swe & fao & got & gsw & avg \\
\midrule 
\multicolumn{3}{l}{\textbf{Baselines}}&\\
\ \ \textsc{mbert} & \multirow{1}{*}{+ \texttt{[T]} (eng)}  & \textbf{69.83} &  \textbf{76.65} &  \textbf{78.27} &  \textbf{89.95} &  \textbf{73.90} & 56.86 & \textbf{79.49} &  \textbf{81.47} &  \textbf{83.09} & 72.80 &   28.20 &  56.43 & 70.58\\
\ \ \textsc{mbert} & \multirow{1}{*}{+ \texttt{[LT]} (eng)}  &  67.97 &  75.56 &  76.89 &  89.28 &  72.22 & 56.65 & 77.79 &  80.07 &  81.72 &  66.93 &   30.15 &  55.23 & 69.20\\
\multicolumn{3}{l}{\textbf{Phylogenically inspired}}&\\
\ \ \textsc{mbert} & \multirow{1}{*}{+ \texttt{[FGLT]} (eng)}  & 68.34 &  76.26 &  77.13 &  89.56 &  73.51 & 61.50 & 78.64 &  80.30 &  81.87 & \textbf{75.70} &   \textbf{46.61} &  \textbf{57.94} & \textbf{72.28}\\
\multicolumn{3}{l}{\textbf{Ablations}}&\\
\ \ \textsc{mbert} & \multirow{1}{*}{+ \texttt{[LT]} (eng)}  & 63.41 &  69.39 &  71.22 &  79.97 &  63.77 & 56.51 & 72.11 &  72.03 &  75.03 & 64.85 &   38.69 &  50.32 & 64.78\\
\ \ \textsc{mbert} & \multirow{1}{*}{+ \texttt{[FLT]} (eng)}  & 68.26 &  76.10 &  77.47 &  89.38 &  73.10 & \textbf{62.40} & 78.52 &  80.39 &  82.12 & 75.05 &   46.02 &  57.81 & 72.22\\
\bottomrule
\end{tabular}

 \begin{tabular}{@{ }l@{}l@{ }|c@{ \ }c@{ \ }c@{ \ }|c@{ \ }c@{ \ }c@{ \ }c@{ \ }c@{ \ }c@{ \ }c@{ \ }c@{ \ }|c@{}}
\multicolumn{14}{c}{\textbf{Uralic}} \\\hline
&&\multicolumn{3}{c}{\textbf{\textsc{mbert-seen}}} & \multicolumn{8}{c}{\textbf{\textsc{mbert-unseen}}}&\\
Model&Training& est &     fin &     hun &    koi &    kpv &    krl &    mdf &    myv &    olo &    sme &    sms &    avg \\
\midrule
\multicolumn{3}{l}{\textbf{Baselines}}&\\
 \ \ \textsc{mbert} & \multirow{1}{*}{+ \texttt{[T]} (est)}  &  83.67 &  78.51 &  73.42 &  29.08 &  24.15 &  55.53 &  40.89 &  36.45 &  56.65 &  29.34 &  23.37 &  48.28 \\
  \ \ \textsc{mbert} & \multirow{1}{*}{+ \texttt{[LT]} (est)}  &  83.95 &  79.41 &  73.10 &  34.68 &  30.87 &  63.41 &  39.23 &  37.58 &  63.10 &  31.85 &  \textbf{28.18} &  51.40 \\
 \multicolumn{3}{l}{\textbf{Phylogenically inspired}}&\\
  \ \ \textsc{mbert} & \multirow{1}{*}{+ \texttt{[FGLT]} (est)}  &  \textbf{84.10} &  \textbf{79.45} &  73.73 &  \textbf{44.10} &  \textbf{42.40} &  \textbf{69.65} &  \textbf{55.88} &  \textbf{53.77} &  \textbf{69.62} &  \textbf{45.16} &  23.00 &  \textbf{58.26} \\
 \multicolumn{3}{l}{\textbf{Ablations}}&\\
 \ \ \textsc{mbert} & \multirow{1}{*}{+ \texttt{[LT]} (est)}  &  75.68 &  71.45 &  66.97 &  36.83 &  32.51 &  60.60 &  41.28 &  39.57 &  62.70 &  33.12 &  23.89 &  49.51 \\
\ \ \textsc{mbert} & \multirow{1}{*}{+ \texttt{[FLT]} (est)}  &   83.72 &  78.84 &  \textbf{73.78} &  37.31 &  34.55 &  68.13 &  50.13 &  47.24 &  68.95 &  41.71 &  24.63 &  55.36 \\
\bottomrule
\end{tabular}
\begin{tabular}{@{ }l@{}l@{ }|c@{ \ }c@{ \ }c@{ \ }c@{ \ }c@{ \ }c@{ \ }c@{ \ }c@{ \ }|c@{ \ }}
\multicolumn{11}{c}{\textbf{Tupian}} \\\hline
&& \multicolumn{8}{c}{\textbf{\textsc{mbert-unseen}}}&\\
 Model & Training & aqz &    arr &    gub &   gun &    mpu &    myu &    tpn &    urb &    avg \\
\midrule 
\multicolumn{3}{l}{\textbf{Baselines}}&\\
\ \ \textsc{mbert} & \multirow{1}{*}{+ \texttt{[T]} (eng)} & \textbf{27.50} &  \textbf{33.82} &  26.07 &   9.11 &  \textbf{23.97} &  \textbf{25.46} &  22.37 &  24.59 &  \textbf{24.11} \\
\ \ \textsc{mbert} & \multirow{1}{*}{+ \texttt{[LT]} (eng)} & 22.50 &  26.66 &  19.69 &  \textbf{11.55} &  17.81 &  19.19 &   9.21 &  25.41 &  19.00 \\
 \multicolumn{3}{l}{\textbf{Phylogenically inspired}}&\\
 \ \ \textsc{mbert} & \multirow{1}{*}{+ \texttt{[FGLT]} (eng)} & \textbf{27.50} &  26.01 &  \textbf{28.46} &  10.45 &  19.86 &  19.56 &  \textbf{31.58} &  \textbf{26.78} &  23.77 \\
 \multicolumn{3}{l}{\textbf{Ablations}}&\\
\ \ \textsc{mbert} & \multirow{1}{*}{+ \texttt{[LT]} (eng)} & 21.25 &  24.20 &  23.78 &  10.30 &  15.75 &  23.62 &  18.42 &  26.50 &  20.48 \\
\ \ \textsc{mbert} & \multirow{1}{*}{+ \texttt{[FLT]} (eng)} & 25.00 &  26.45 &  26.66 &   9.86 &  17.12 &  20.30 &  19.74 &  22.68 &  20.97 \\

\bottomrule
     \end{tabular}
   \caption{Dependency Parsing Task Results (base model:  \textsc{mbert}, metric: UAS).}
    \label{tab:test_dep}
\end{table*}


\begin{table*}[!t]
    \centering
    \small
    \begin{tabular}{@{ }l@{}l@{ }|c@{ \ }c@{ \ }c@{ \ }c@{ \ }c@{ \ }c@{ \ }c@{ \ }c@{ \ }c@{ \ }|c@{ \ }c@{ \ }c@{ \ }|c@{}}

\toprule
\multicolumn{15}{c}{\textbf{Germanic}} \\
\midrule
&&\multicolumn{9}{c}{\textbf{\textsc{xlm-r-seen}}} & \multicolumn{3}{c}{\textbf{\textsc{xlm-r-unseen}}}&\\
Model&Training& afr & dan & deu & eng & isl & nds & nld & nor & swe & fao & got & gsw & avg \\
\midrule 
\multicolumn{3}{l}{\textbf{Baselines}}&\\
\ \ \textsc{xlm-r} & \multirow{1}{*}{+ \texttt{[T]} (eng)}  & 68.36 &  74.82 &  77.07 &  85.00 &  74.36 &  44.73 &  77.01 &  79.66 &  81.94 &  70.20 &  25.04 &  42.87 &  66.75 \\
\ \ \textsc{xlm-r} & \multirow{1}{*}{+ \texttt{[LT]} (eng)}  &  \textbf{69.78} &  76.38 & \textbf{ 78.54} &  87.22 &  \textbf{76.12} &  56.60 &  \textbf{78.70} &  \textbf{81.43} &  \textbf{83.46} &  \textbf{74.17} &  23.47 &  56.37 &  70.19 \\
\multicolumn{3}{l}{\textbf{Phylogenically inspired}}&\\
\ \ \textsc{xlm-r} & \multirow{1}{*}{+ \texttt{[FGLT]} (eng)} & 69.74 &  \textbf{76.56} &  78.00 & \textbf{ 87.38} &  75.80 &  \textbf{58.54} &  78.68 &  81.33 &  83.31 &  73.47 &  \textbf{38.18} &  \textbf{63.09} &  \textbf{72.01} \\
\multicolumn{3}{l}{\textbf{Ablations}}&\\
\ \ \textsc{xlm-r} & \multirow{1}{*}{+ \texttt{[LT]} (eng)}  & 67.67 &  73.73 &  75.52 &  83.65 &  73.30 &  53.16 &  76.33 &  78.65 &  80.86 &  68.68 &  32.45 &  55.40 &  68.28 \\
\ \ \textsc{xlm-r} & \multirow{1}{*}{+ \texttt{[FLT]} (eng)}  & 69.66 &  76.41 &  78.11 &  87.29 &  75.97 &  57.63 &  78.75 &  81.49 &  83.44 &  73.67 &  36.88 &  62.53 &  71.82 \\
\bottomrule
\end{tabular}

 \begin{tabular}{@{ }l@{}l@{ }|c@{ \ }c@{ \ }c@{ \ }|c@{ \ }c@{ \ }c@{ \ }c@{ \ }c@{ \ }c@{ \ }c@{ \ }c@{ \ }|c@{}}
\multicolumn{14}{c}{\textbf{Uralic}} \\\hline
&&\multicolumn{3}{c}{\textbf{\textsc{xlm-r-seen}}} & \multicolumn{8}{c}{\textbf{\textsc{xlm-r-unseen}}}&\\
Model&Training& est &     fin &     hun &    koi &    kpv &    krl &    mdf &    myv &    olo &    sme &    sms &    avg \\
\midrule
\multicolumn{3}{l}{\textbf{Baselines}}&\\
 \ \ \textsc{xlm-r} & \multirow{1}{*}{+ \texttt{[T]} (est)}  &  82.02 &  78.59 &  73.16 &  31.94 &  30.25 &  61.47 &  34.41 &  34.46 &  56.45 &  26.27 &  31.07 &  49.10 \\
  \ \ \textsc{xlm-r} & \multirow{1}{*}{+ \texttt{[LT]} (est)}  &  \textbf{84.25} &  \textbf{80.11} &  \textbf{74.72} &  33.37 &  31.31 &  65.03 &  33.62 &  31.91 &  58.47 &  25.72 &  28.25 &  49.71 \\
 \multicolumn{3}{l}{\textbf{Phylogenically inspired}}&\\
  \ \ \textsc{xlm-r} & \multirow{1}{*}{+ \texttt{[FGLT]} (est)} & 83.39 &  79.40 &  73.61 &  \textbf{40.76} &  \textbf{39.00} &  \textbf{67.84} &  \textbf{37.71} &  \textbf{38.66} &  \textbf{67.07} &  \textbf{29.11} &  31.21 &  \textbf{53.44} \\
 \multicolumn{3}{l}{\textbf{Ablations}}&\\
 \ \ \textsc{xlm-r} & \multirow{1}{*}{+ \texttt{[LT]} (est)}  &  81.67 &  77.80 &  72.14 &  33.85 &  30.71 &  62.57 &  30.18 &  33.44 &  63.44 &  23.96 &  30.33 &  49.10 \\
\ \ \textsc{xlm-r} & \multirow{1}{*}{+ \texttt{[FLT]} (est)} &  83.22 &  79.41 &  74.05 &  39.93 &  38.12 &  66.52 &  37.25 &  38.20 &  66.20 &  28.23 & \textbf{ 31.73} &  52.99 \\
\bottomrule
\end{tabular}
\begin{tabular}{@{ }l@{}l@{ }|c@{ \ }c@{ \ }c@{ \ }c@{ \ }c@{ \ }c@{ \ }c@{ \ }c@{ \ }|c@{ \ }}
\multicolumn{11}{c}{\textbf{Tupian}} \\\hline
&& \multicolumn{8}{c}{\textbf{\textsc{xlm-r-unseen}}}&\\
 Model & Training & aqz &    arr &    gub &   gun &    mpu &    myu &    tpn &    urb &    avg \\
\midrule 
\multicolumn{3}{l}{\textbf{Baselines}}&\\
\ \ \textsc{xlm-r} & \multirow{1}{*}{+ \texttt{[T]} (eng)} & \textbf{33.75} &  \textbf{29.47} &  17.40 &  3.95 &  \textbf{24.66} &  \textbf{30.63} &  19.74 &  25.14 &  23.09 \\
\ \ \textsc{xlm-r} & \multirow{1}{*}{+ \texttt{[LT]} (eng)} & 32.50 &  28.99 &  17.88 &  \textbf{3.96} &  21.92 &  27.68 &  22.37 &  24.86 &  22.52 \\
 \multicolumn{3}{l}{\textbf{Phylogenically inspired}}&\\
 \ \ \textsc{xlm-r} & \multirow{1}{*}{+ \texttt{[FGLT]} (eng)} & 27.50 &  28.52 &  \textbf{28.51} &  3.84 &  23.29 &  28.41 &  25.00 &  \textbf{28.69} &  \textbf{24.22} \\
 \multicolumn{3}{l}{\textbf{Ablations}}&\\
\ \ \textsc{xlm-r} & \multirow{1}{*}{+ \texttt{[LT]} (eng)} & 27.50 &  29.25 &  19.40 &  3.38 &  21.23 &  26.57 &  \textbf{28.95} &  19.40 &  21.96 \\
\ \ \textsc{xlm-r} & \multirow{1}{*}{+ \texttt{[FLT]} (eng)} & 23.75 &  28.82 &  23.59 &  3.50 &  19.86 &  28.04 &  23.68 &  26.50 &  22.22 \\

\bottomrule
     \end{tabular}
    \caption{Dependency Parsing Task Results (base model:  \textsc{xlm-r}, metric: UAS).}
    \label{tab:test_dep_xlmr}
\end{table*}

\paragraph{POS Tagging}
For POS tagging task, we select the same language and settings like before we used in dependency parsing. In POS tagging, the language adapter does not make negative interference like it made in case of dependency parsing. However, using phylogny based joint training still performs better than all the baseline in majority Germanic and Uralic languages. In case of Tupian languages, we see improvement using phylogeny based adaptation in 4 out of 8 languages.  
\begin{table*}[!t]
    \centering
    \small
    \begin{tabular}{@{ }l@{}l@{ }|c@{ \ }c@{ \ }c@{ \ }c@{ \ }c@{ \ }c@{ \ }c@{ \ }c@{ \ }c@{ \ }|c@{ \ }c@{ \ }c@{ \ }|c@{}}

\toprule
\multicolumn{15}{c}{\textbf{Germanic}} \\
\midrule
&&\multicolumn{9}{c}{\textbf{\textsc{mbert-seen}}} & \multicolumn{3}{c}{\textbf{\textsc{mbert-unseen}}}&\\
Model&Training& afr & dan & deu & eng & isl & nds & nld & nor & swe & fao & got & gsw & avg \\
\midrule 
\multicolumn{3}{l}{\textbf{Baselines}}&\\
\ \ \textsc{mbert} & \multirow{1}{*}{+ \texttt{[T]} (eng)}  & 85.08 &  87.55 &  85.04 &  95.50 &  83.18 &  69.53 &  87.88 &  90.49 &  89.74 &  80.70 &   14.50 &  58.18 &  77.28 \\
\ \ \textsc{mbert} & \multirow{1}{*}{+ \texttt{[LT]} (eng)}  &  85.93 &  88.23 &  86.16 &  95.64 &  84.49 &  72.93 &  87.70 &  90.22 &  90.10 &  79.93 &   22.60 &  71.07 &  79.58 \\
\multicolumn{3}{l}{\textbf{Phylogenically inspired}}&\\
\ \ \textsc{mbert} & \multirow{1}{*}{+ \texttt{[FGLT]} (eng)}  & \textbf{86.09} &  88.31 &  \textbf{86.27} &  \textbf{95.66} &  \textbf{84.83} &  \textbf{74.54} &  88.06 &  90.50 &  \textbf{90.10} &  \textbf{88.88} &   \textbf{56.03} &  \textbf{74.86} &  \textbf{83.68} \\
\multicolumn{3}{l}{\textbf{Ablations}}&\\
\ \ \textsc{mbert} & \multirow{1}{*}{+ \texttt{[LT]} (eng)}  & 85.03 &  87.40 &  84.68 &  94.23 &  82.89 &  71.82 &  86.37 &  88.18 &  88.61 &  82.31 &   47.23 &  70.25 &  80.75 \\
\ \ \textsc{mbert} & \multirow{1}{*}{+ \texttt{[FLT]} (eng)}  & 86.08 &  \textbf{88.36} &  86.08 &  95.62 &  84.45 &  73.86 &  \textbf{88.15} &  \textbf{90.52} &  89.95 &  88.15 &   55.47 &  73.65 &  83.36 \\
\bottomrule
\end{tabular}

 \begin{tabular}{@{ }l@{}l@{ }|c@{ \ }c@{ \ }c@{ \ }|c@{ \ }c@{ \ }c@{ \ }c@{ \ }c@{ \ }c@{ \ }c@{ \ }c@{ \ }|c@{}}
\multicolumn{14}{c}{\textbf{Uralic}} \\\hline
&&\multicolumn{3}{c}{\textbf{\textsc{mbert-seen}}} & \multicolumn{8}{c}{\textbf{\textsc{mbert-unseen}}}&\\
Model&Training& est &     fin &     hun &    koi &    kpv &    krl &    mdf &    myv &    olo &    sme &    sms & avg \\
\midrule
\multicolumn{3}{l}{\textbf{Baselines}}&\\
 \ \ \textsc{mbert} & \multirow{1}{*}{+ \texttt{[T]} (est)}  &  89.39 &  82.85 &  70.07 &  32.22 &  24.02 &  62.79 &  46.53 &  43.79 &  62.67 &  40.15 &  23.21 &  52.52 \\
  \ \ \textsc{mbert} & \multirow{1}{*}{+ \texttt{[LT]} (est)}  &  89.49 &  83.29 &  70.38 &  46.78 &  35.96 &  70.78 &  46.55 &  41.26 &  65.37 &  44.46 &  \textbf{29.03} &  56.67 \\
 \multicolumn{3}{l}{\textbf{Phylogenically inspired}}&\\
  \ \ \textsc{mbert} & \multirow{1}{*}{+ \texttt{[FGLT]} (est)}  &  \textbf{90.88} &  \textbf{84.93} &  69.98 &  \textbf{49.01} &  \textbf{41.74} &  \textbf{79.17} &  \textbf{60.69} &  \textbf{57.69} &  \textbf{73.75} &  \textbf{55.27} &  20.32 &  \textbf{62.13} \\
 \multicolumn{3}{l}{\textbf{Ablations}}&\\
 \ \ \textsc{mbert} & \multirow{1}{*}{+ \texttt{[LT]} (est)}  &  87.12 &  82.21 &  68.67 &  39.83 &  34.73 &  72.90 &  50.58 &  45.83 &  67.80 &  49.13 &  25.44 &  56.75 \\
\ \ \textsc{mbert} & \multirow{1}{*}{+ \texttt{[FLT]} (est)}  &   90.55 &  83.99 &  \textbf{70.45} &  41.96 &  36.64 &  76.76 &  52.89 &  50.25 &  70.62 &  51.28 &  20.56 &  58.72 \\

\bottomrule 
\end{tabular}
\begin{tabular}{@{ }l@{}l@{ }|c@{ \ }c@{ \ }c@{ \ }c@{ \ }c@{ \ }c@{ \ }c@{ \ }c@{ \ }|c@{ \ }}

\multicolumn{11}{c}{\textbf{Tupian}} \\\hline
&& \multicolumn{8}{c}{\textbf{\textsc{mbert-unseen}}}&\\
 Model & Training & aqz &   arr &    gub &   gun &   mpu &    myu &    tpn &    urb &    avg \\
 \midrule
\multicolumn{3}{l}{\textbf{Baselines}}&\\
\ \ \textsc{mbert-r} & \multirow{1}{*}{+ \texttt{[T]} (eng)} & 9.60 &  3.06 &  23.02 &  0.37 &  4.95 &  15.52 &  18.02 &   4.79 &   9.92 \\
\ \ \textsc{mbert-r} & \multirow{1}{*}{+ \texttt{[LT]} (eng)} & \textbf{19.35} &  4.88 &  26.21 &  \textbf{2.42} &  \textbf{6.25} &  19.33 &  20.00 &   7.13 &  \textbf{13.20} \\
 \multicolumn{3}{l}{\textbf{Phylogenically inspired}}&\\
 \ \ \textsc{mbert-r} & \multirow{1}{*}{+ \texttt{[FGLT]} (eng)} & 12.28 &  \textbf{5.44} &  26.32 &  0.23 &  5.62 &  19.49 &  \textbf{24.39} &   7.10 &  12.61 \\
 \multicolumn{3}{l}{\textbf{Ablations}}&\\
\ \ \textsc{mbert-r} & \multirow{1}{*}{+ \texttt{[LT]} (eng)} & 13.79 &  3.65 &  \textbf{26.92} &  0.21 &  3.57 &  17.37 &  17.86 &   6.60 &  11.25 \\
\ \ \textsc{mbert-r} & \multirow{1}{*}{+ \texttt{[FLT]} (eng)} & 18.64 &  3.71 &  26.62 &  0.20 &  4.68 &  \textbf{21.01} &  21.31 &   \textbf{7.43} &  12.95 \\

\bottomrule
     \end{tabular}
     \caption{Parts of Speech Task Results (base model:  \textsc{mbert}, metric: F1).}
    \label{tab:test_pos}
\end{table*}


\begin{table*}[!t]
    \centering
    \small
    \begin{tabular}{@{ }l@{}l@{ }|c@{ \ }c@{ \ }c@{ \ }c@{ \ }c@{ \ }c@{ \ }c@{ \ }c@{ \ }c@{ \ }|c@{ \ }c@{ \ }c@{ \ }|c@{}}

\toprule
\multicolumn{15}{c}{\textbf{Germanic}} \\
\midrule
&&\multicolumn{9}{c}{\textbf{\textsc{xlm-r-seen}}} & \multicolumn{3}{c}{\textbf{\textsc{xlm-r-unseen}}}&\\
Model&Training& afr & dan & deu & eng & isl & nds & nld & nor & swe & fao & got & gsw & avg \\
\midrule 
\multicolumn{3}{l}{\textbf{Baselines}}&\\
\ \ \textsc{xlm-r} & \multirow{1}{*}{+ \texttt{[T]} (eng)}  & \textbf{87.27} &  \textbf{89.14} &  87.64 &  96.34 &  85.64 &  55.77 &  87.75 &  91.15 &  91.49 &  81.29 &  16.50 &  47.67 &  76.47 \\
\ \ \textsc{xlm-r} & \multirow{1}{*}{+ \texttt{[LT]} (eng)}  &  87.25 &  89.05 &  87.53 &  96.36 &  85.55 &  70.21 &  87.73 &  91.12 &  91.35 &  87.16 &  15.41 &  66.37 &  79.59 \\
\multicolumn{3}{l}{\textbf{Phylogenically inspired}}&\\
\ \ \textsc{xlm-r} & \multirow{1}{*}{+ \texttt{[FGLT]} (eng)} & 86.98 &  88.94 &  \textbf{88.09} &  \textbf{96.44} &  85.62 &  \textbf{74.31} &  \textbf{87.94} &  91.11 &  91.35 &  \textbf{88.85} &  \textbf{41.75} &  \textbf{76.52} &  \textbf{83.16} \\
\multicolumn{3}{l}{\textbf{Ablations}}&\\
\ \ \textsc{xlm-r} & \multirow{1}{*}{+ \texttt{[LT]} (eng)}  & 86.75 &  89.05 &  87.77 &  96.36 &  \textbf{85.80} &  71.16 &  87.89 &  91.08 & \textbf{ 91.52} &  88.23 &  34.60 &  68.65 &  81.57 \\
\ \ \textsc{xlm-r} & \multirow{1}{*}{+ \texttt{[FLT]} (eng)}  & 86.92 &  89.00 &  87.86 &  96.40 &  85.78 &  72.39 &  87.97 &  \textbf{91.17} &  91.38 &  88.81 &  39.23 &  73.43 &  82.53 \\
\bottomrule
\end{tabular}

 \begin{tabular}{@{ }l@{}l@{ }|c@{ \ }c@{ \ }c@{ \ }|c@{ \ }c@{ \ }c@{ \ }c@{ \ }c@{ \ }c@{ \ }c@{ \ }c@{ \ }|c@{}}
\multicolumn{14}{c}{\textbf{Uralic}} \\\hline
&&\multicolumn{3}{c}{\textbf{\textsc{xlm-r-seen}}} & \multicolumn{8}{c}{\textbf{\textsc{xlm-r-unseen}}}&\\
Model&Training& est &     fin &     hun &    koi &    kpv &    krl &    mdf &    myv &    olo &    sme &    sms & avg \\
\midrule
\multicolumn{3}{l}{\textbf{Baselines}}&\\
 \ \ \textsc{xlm-r} & \multirow{1}{*}{+ \texttt{[T]} (est)}  &  96.61 &  \textbf{89.31} &  83.98 &  47.30 &  38.39 &  70.39 &  43.15 &  44.21 &  64.99 &  37.74 &  34.84 &  59.17 \\
  \ \ \textsc{xlm-r} & \multirow{1}{*}{+ \texttt{[LT]} (est)}  &  96.64 &  89.30 &  83.61 &  46.97 &  39.57 &  74.55 &  41.89 &  43.95 &  65.86 &  36.58 &  33.32 &  59.29 \\
 \multicolumn{3}{l}{\textbf{Phylogenically inspired}}&\\
  \ \ \textsc{xlm-r} & \multirow{1}{*}{+ \texttt{[FGLT]} (est)}  & 96.69 &  89.23 &  83.31 &  \textbf{56.93} &  \textbf{47.37} &  \textbf{81.41} &  \textbf{47.88} &  \textbf{49.40} &  \textbf{73.71} &  \textbf{46.68} &  \textbf{35.79} &  \textbf{64.40} \\
 \multicolumn{3}{l}{\textbf{Ablations}}&\\
 \ \ \textsc{xlm-r} & \multirow{1}{*}{+ \texttt{[LT]} (est)}  &  96.54 &  89.22 &  83.61 &  48.42 &  41.07 &  80.00 &  43.87 &  46.01 &  72.15 &  41.63 &  35.15 &  61.61 \\
\ \ \textsc{xlm-r} & \multirow{1}{*}{+ \texttt{[FLT]} (est)}  &   \textbf{96.71} &  89.21 &  \textbf{84.24} &  50.38 &  42.94 &  80.70 &  44.88 &  46.29 &  72.71 &  42.05 &  35.96 &  62.37 \\
\bottomrule
\end{tabular}
\begin{tabular}{@{ }l@{}l@{ }|c@{ \ }c@{ \ }c@{ \ }c@{ \ }c@{ \ }c@{ \ }c@{ \ }c@{ \ }|c@{ \ }}

\multicolumn{11}{c}{\textbf{Tupian}} \\\hline
&& \multicolumn{8}{c}{\textbf{\textsc{xlm-r-unseen}}}&\\
 Model & Training & aqz &   arr &    gub &   gun &   mpu &    myu &    tpn &    urb &    avg \\
\midrule 
\multicolumn{3}{l}{\textbf{Baselines}}&\\
\ \ \textsc{xlm-r-r} & \multirow{1}{*}{+ \texttt{[T]} (eng)} & 6.25 &  \textbf{5.92} &  26.05 &  \textbf{5.13} &  8.16 &  16.07 &  21.62 &   6.91 &  12.01 \\
\ \ \textsc{xlm-r-r} & \multirow{1}{*}{+ \texttt{[LT]} (eng)} &   6.96 &  4.80 &  27.16 &  2.67 &  6.10 &  20.96 &  26.79 &   6.56 &  12.75 \\
 \multicolumn{3}{l}{\textbf{Phylogenically inspired}}&\\
 \ \ \textsc{xlm-r-r} & \multirow{1}{*}{+ \texttt{[FGLT]} (eng)} & 11.86 &  4.89 &  \textbf{37.35} &  4.35 &  7.27 &  \textbf{23.86} &  23.53 &  \textbf{12.74} &  \textbf{15.73} \\
 \multicolumn{3}{l}{\textbf{Ablations}}&\\
\ \ \textsc{xlm-r-r} & \multirow{1}{*}{+ \texttt{[LT]} (eng)} & \textbf{15.83} &  5.36 &  27.05 &  4.26 &  \textbf{9.85} &  13.91 &  \textbf{26.67} &   8.11 &  13.88 \\
\ \ \textsc{xlm-r-r} & \multirow{1}{*}{+ \texttt{[FLT]} (eng)} & 12.60 &  4.36 &  32.19 &  4.58 &  4.52 &  17.53 &  25.64 &   8.98 &  13.80 \\

\bottomrule
     \end{tabular}
    \caption{Parts of Speech Task Results (base model:  \textsc{xlm-r}, metric: F1).}
    \label{tab:test_pos_xlmr}
\end{table*}

\paragraph{NLI}
Our NLI results are presented in table \ref{tab:test_nli_mbert} and \ref{tab:test_nli}. In addition, we reprot the zero-shot baseline results from \cite{DBLP:journals/corr/abs-2104-08726} where the pretrained language model was continually trained on monolingual task language before training on downstream english task data. In our adaptation settings, we follow the \texttt{[FGLT]} combinations. Our approach does better for low resource ones (i.e.) while joint training results in optimal performance. 
\begin{table}[t]
\small
    \centering
    \begin{tabular}{@{ }l@{}l@{ }|c@{ \ }c@{ \ }c@{ \ }c@{ \ }|c@{}}

\toprule
 Model & Training &  grn & hch & nah &  tar & avg \\
\midrule 
\multicolumn{4}{l}{\textbf{Baselines}} \\
\ \ \textsc{mbert} & \multirow{1}{*}{+ \texttt{[T]} (eng)}  &  33.60 &  33.20 &  33.60 &  33.33 &  33.43 \\
\ \ \textsc{mbert} & \multirow{1}{*}{+ \texttt{[LT]} (eng)}  & 34.40 &  33.20 &  33.60 &  \textbf{33.73} &  33.73 \\
\midrule
\multicolumn{4}{l}{\textbf{Phylogenically inspired}} \\
\ \ \textsc{mbert} & \multirow{1}{*}{+ \texttt{[FGLT]} (eng)} & \textbf{36.13} &  33.47 & \textbf{ 33.88 }&  33.33 &  \textbf{34.20} \\
\multicolumn{4}{l}{\textbf{Ablations}} \\
\ \ \textsc{mbert} & \multirow{1}{*}{+ \texttt{[LT]} (eng)} & 33.33 &  33.33 &  33.20 &  33.07 &  33.23 \\
\ \ \textsc{mbert} & \multirow{1}{*}{+ \texttt{[FLT]} (eng)}   &  \textbf{33.73} &  33.73 &  33.47 &  33.33 &  33.57 \\
\bottomrule
     \end{tabular}
    \caption{NLI Task Results on AmericasNLI~\cite{DBLP:journals/corr/abs-2104-08726} languages (base model:  \textsc{mbert}, metric: ACC).}
    \label{tab:test_nli_mbert}
\end{table}

\begin{table}[t]
\small
    \centering
    \begin{tabular}{@{ }l@{}l@{ }|c@{ \ }c@{ \ }c@{ \ }c@{ \ }|c@{}}

\toprule
 Model & Training &  grn & hch & nah &  tar & avg \\
\midrule 
\multicolumn{4}{l}{\textbf{Baselines}} \\
\ \ \textsc{xlm-r} & \multirow{1}{*}{+ \texttt{[T]} (eng)}  &  45.33 &  38.27 &  42.01  &  38.40 &  41.00 \\
\ \ \textsc{xlm-r} & \multirow{1}{*}{+ \texttt{[LT]} (eng)}  &  44.40 &  \textbf{38.53} &  \textbf{47.83} &  37.47 &  42.06 \\
\midrule
\multicolumn{4}{l}{\textbf{Phylogenically inspired}} \\
\ \ \textsc{xlm-r} & \multirow{1}{*}{+ \texttt{[FGLT]} (eng)} & 46.27 &  37.60 &  47.15 &  \textbf{40.67} &  \textbf{42.92} \\
\multicolumn{4}{l}{\textbf{Ablations}} \\
\ \ \textsc{xlm-r} & \multirow{1}{*}{+ \texttt{[LT]} (eng)} & 46.27 &  37.20 &  44.17  &  40.27 &  41.98 \\
\ \ \textsc{xlm-r} & \multirow{1}{*}{+ \texttt{[FLT]} (eng)}   &  \textbf{47.87} &  38.27 &  45.66 &  38.27 &  42.52 \\
\midrule
\multicolumn{4}{l}{\textbf{zero shot w/ mlm baseline}:} \\
\ \ \textsc{xlm-r} & \multirow{1}{*}{+mlm (eng)}  & 52.44 & 37.25 & 46.21 & 39.82 & 43.93  \\ 
\bottomrule
     \end{tabular}
    \caption{NLI Task Results on AmericasNLI~\cite{DBLP:journals/corr/abs-2104-08726} languages (base model:  \textsc{xlm-r}, metric: ACC).}
    \label{tab:test_nli}
\end{table}

\section{Dependency Parsing on Indo-European Family}
The dependency parsing results comprising Indo-European family branches are presented in table \ref{tab:ie_mbert_dep}.
\begin{table*}[!t]
    \centering
    \small
    \begin{tabular}{@{ }l@{}l@{ }|c@{ \ }c@{ \ }c@{ \ }c@{ \ }c@{ \ }|c@{ \ }}

\toprule
\multicolumn{8}{c}{\textbf{Celtic}} \\
\midrule
Model&Training& bre &         wel &         gle &         gla &        glv &        avg \\
\midrule 
 \textsc{mbert} & \multirow{1}{*}{+ \texttt{[FGLT]} (gle)}  & 23.48 &  23.17 &  27.60 &  20.60 &  13.84 &  21.74 \\
  \textsc{mbert} & \multirow{1}{*}{+ \texttt{[RFGLT]} (gle)}  & 17.63 &  21.32 &  28.40 &  17.92 &   9.08 &  18.87 \\
\bottomrule
     \end{tabular}

\begin{tabular}{@{ }l@{}l@{ }|c@{ \ }c@{ \ }c@{ \ }c@{ \ }c@{ \ }c@{ \ }c@{ \ }c@{ \ }c@{ \ }c@{ \ }c@{ \ }c@{ \ }|c@{ \ }}

\multicolumn{15}{c}{\textbf{Germanic}} \\
\midrule
Model&Training& afr &         dan &         deu &         eng &        fao &        got &        gsw &         isl &        nds &         nld &         nor &         swe &        avg \\
\midrule 
 \textsc{mbert} & \multirow{1}{*}{+ \texttt{[FGLT]} (eng)}  & 69.18 &  76.51 &  77.79 &  90.34 &  76.86 &  48.28 &  65.30 &  73.25 &  54.88 &  78.86 &  81.20 &  82.59 &  72.92 \\
  \textsc{mbert} & \multirow{1}{*}{+ \texttt{[RFGLT]} (eng)}  & 63.79 &  70.82 &  70.75 &  84.52 &  65.79 &  41.63 &  53.81 &  66.55 &  49.59 &  70.98 &  73.99 &  76.07 &  65.69 \\
\bottomrule
     \end{tabular}

\begin{tabular}{@{ }l@{}l@{ }|c@{ \ }c@{ \ }c@{ \ }c@{ \ }c@{ \ }c@{ \ }c@{ \ }|c@{ \ }}
\multicolumn{10}{c}{\textbf{Indic}} \\
\midrule
Model&Training& bho &     ben &     hin &     mar &     san &     urd &    xnr &    avg \\
\midrule 
 \textsc{mbert} & \multirow{1}{*}{+ \texttt{[FGLT]} (mar)}  & 16.61 &  54.69 &  19.55 &  58.25 &  23.67 &  14.72 &  32.42 &  31.42 \\
  \textsc{mbert} & \multirow{1}{*}{+ \texttt{[RFGLT]} (mar)}  & 18.50 &  31.25 &  18.55 &  49.76 &  17.42 &  10.61 &  30.63 &  25.24 \\
\bottomrule
     \end{tabular}

\begin{tabular}{@{ }l@{}l@{ }|c@{ \ }c@{ \ }|c@{ \ }}
\multicolumn{5}{c}{\textbf{Iranian}} \\
\midrule
Model&Training& fas &    kmr &    avg \\
\midrule 
 \textsc{mbert} & \multirow{1}{*}{+ \texttt{[FGLT]} (fas)}  &  91.07 &  41.64 &  66.35 \\
  \textsc{mbert} & \multirow{1}{*}{+ \texttt{[RFGLT]} (fas)}  & 86.02 &  36.95 &  61.49 \\
\bottomrule
     \end{tabular}

\begin{tabular}{@{ }l@{}l@{ }|c@{ \ }c@{ \ }c@{ \ }c@{ \ }c@{ \ }c@{ \ }c@{ \ }c@{ \ }c@{ \ }c@{ \ }|c@{ \ }}

\multicolumn{13}{c}{\textbf{Romance}} \\
\midrule
Model&Training& cat &     spa &     fre &    fro &     glg &     ita &    lig &   nap &     por &     rum &    avg \\
\midrule 
 \textsc{mbert} & \multirow{1}{*}{+ \texttt{[FGLT]} (spa)}  & 90.63 &  92.44 &  84.25 &  58.09 &  74.74 &  82.24 &  68.61 &  70.0 &  86.05 &  82.84 &  78.99 \\
  \textsc{mbert} & \multirow{1}{*}{+ \texttt{[RFGLT]} (spa)}  & 80.50 &  82.04 &  72.94 &  42.40 &  68.76 &  71.60 &  58.98 &  50.0 &  73.48 &  68.79 &  66.95 \\
\bottomrule
     \end{tabular} 
     
\begin{tabular}{@{ }l@{}l@{ }|c@{ \ }c@{ \ }c@{ \ }c@{ \ }c@{ \ }c@{ \ }c@{ \ }c@{ \ }c@{ \ }c@{ \ }c@{ \ }c@{ \ }|c@{ \ }}

\multicolumn{15}{c}{\textbf{Slavic}} \\
\midrule
Model&Training&  bel &     bul &    chu &     ces &     hrv &    orv &     pol &  qpm &     rus &     slk &     slv &     srp &  avg \\
\midrule 
 \textsc{mbert} & \multirow{1}{*}{+ \texttt{[FGLT]} (rus)}  & 77.28 &  79.98 &  32.25 &  78.35 &  79.17 &  62.26 &  80.39 &  62.57 &  77.83 &  82.07 &  81.48 &  80.31 &  72.83 \\
  \textsc{mbert} & \multirow{1}{*}{+ \texttt{[RFGLT]} (rus)}  & 68.77 &  69.54 &  28.54 &  67.72 &  68.69 &  55.96 &  68.59 &  49.13 &  65.93 &  69.05 &  71.39 &  72.08 &  62.95 \\
\bottomrule
     \end{tabular} 
     
     \caption{Dependency Parsing Task Results on Indo-European language family (base model: \textsc{mbert}, metric: UAS).}
    \label{tab:ie_mbert_dep}
\end{table*}


\section{Random Family Tree}
 \label{app_sec:rand}
In our random family tree construction, we select  9 languages from 9 different language family branches. We group these languages into 3 genus while keeping one language in each genus from either Germanic,  Tupian or Uralic language family on which we report our experimental result. The tree structer is presented in table \ref{tab:rand_tree}.
\begin{table*}[!t]
    \centering
    \small
    \begin{tabular}{llll}
\toprule
      Family &        Genus &              Language (Original Family) &                ISO 639-3  \\
\midrule
 \multirow{11}{*}{Random} &       R1 &                Bulgarian (Slavic) &                bul  \\
  &      R1 &                Irish (Celtic) &                gle  \\
    &      R1 &                Kaapor (Tupian) &                urb  \\\\
    &      R2 &                Basque (Language Isolate) &                baq  \\
    &      R2 &                Komi Zyrian (Uralic) &                kpv  \\
    &      R2 &                Telugu (Dravidian) &                tel  \\\\
    &      R3 &                Faroese (Germanic) &                fao  \\
    &      R3 &                Hebrew (Semitic) &                heb  \\
    &      R3 &                Hindi (Indic) &                hin  \\
\bottomrule
\end{tabular}
    \caption{Random Language Family construction.}
    \label{tab:rand_tree}
\end{table*}



\end{document}